%% file: main.tex
\documentclass[10pt,twocolumn,letterpaper]{article}

\usepackage[pagenumbers]{iccv} 

\input{preamble}

\usepackage{multirow}
\usepackage{pifont}
\usepackage{listings}
\usepackage{float}
\usepackage{xcolor}

\lstdefinestyle{custompython}{
    backgroundcolor=\color{gray!10},   
    basicstyle=\ttfamily\footnotesize, 
    breaklines=true,                    
    frame=tb,                           
    keywordstyle=\color{blue},          
    commentstyle=\color{gray!60!black}, 
    stringstyle=\color{red},
    tabsize=4,
    captionpos=b,                        
    numbers=left,                        
    numberstyle=\tiny\color{gray},       
    xleftmargin=15pt, xrightmargin=10pt, 
    columns=fullflexible                 
}
\definecolor{iccvblue}{rgb}{0.21,0.49,0.74}
\usepackage[pagebackref,breaklinks,colorlinks,allcolors=iccvblue]{hyperref}

\def\paperID{13498} 
\def\confName{ICCV}
\def\confYear{2025}

\usepackage{graphicx}
\usepackage{graphics}
\DeclareGraphicsExtensions{.pdf,.png,.jpg} 
\graphicspath{{./figures/}}
\definecolor{custompink}{RGB}{255,20,144}
\hypersetup{
    colorlinks=true,
    linkcolor=custompink,
    urlcolor=custompink
}

\begin{document}

\title{AIGVE-Tool: AI-Generated Video Evaluation Toolkit \\with Multifaceted Benchmark} 

\author{Xinhao Xiang$^{\ast}$ \quad Xiao Liu$^{\ast}$ \quad Zizhong Li \quad Zhuosheng Liu \quad Jiawei Zhang\\
IFM Lab, University of California, Davis\\
{\tt\small \{xhxiang, xioliu, zzoli, zsliu, jiwzhang\}@ucdavis.edu} \\
Project Website: \href{https://www.aigve.org/}{https://www.aigve.org/} \\
Github: \href{https://github.com/ShaneXiangH/AIGVE_Tool}{https://github.com/ShaneXiangH/AIGVE\_Tool}
}

\maketitle
\renewcommand{\thefootnote}{\fnsymbol{footnote}}
\footnotetext[1]{Equal contribution.}

\input{content/0-abstract}  
\input{content/1-introduction}

\input{content/2-related}

\input{content/3-toolkit}

\input{content/4-benchmark}

\input{content/5-experiment}

\input{content/6-conclusion}



{\small
\bibliographystyle{ieee_fullname}
\bibliography{main}
}

\appendix
\input{content/Appendix}

\end{document}

%% file: preamble.tex
\usepackage{times}
\usepackage{amsmath}
\usepackage{amssymb}
\usepackage{booktabs}
\usepackage{colortbl}
\usepackage{array}
\usepackage{tabularx}


%% file: content/0-abstract.tex
\begin{abstract}
The rapid advancement in AI-generated video synthesis has led to a growth demand for standardized and effective evaluation metrics. Existing metrics lack a unified framework for systematically categorizing methodologies, limiting a holistic understanding of the evaluation landscape. 
Additionally, fragmented implementations and the absence of standardized interfaces lead to redundant processing overhead. 
Furthermore, many prior approaches are constrained by dataset-specific dependencies, limiting their applicability across diverse video domains.
To address these challenges, we introduce \ours (AI-Generated Video Evaluation Toolkit), a unified framework that provides a structured and extensible evaluation pipeline for a comprehensive AI-generated video evaluation. Organized within a novel five-category taxonomy, \ours integrates multiple evaluation methodologies while allowing flexible customization through a modular configuration system. Additionally, we propose \data, a large-scale benchmark dataset created with five SOTA video generation models based on hand-crafted instructions and prompts. This dataset systematically evaluates various video generation models across nine critical quality dimensions. Extensive experiments demonstrate the effectiveness of \ours in providing standardized and reliable evaluation results, highlighting specific strengths and limitations of current models and facilitating the advancements of next-generation AI-generated video techniques.

\end{abstract}

%% file: content/1-introduction.tex
\section{Introduction}
\label{sec:intro}

Recent advances in deep generative models have revolutionized AI-Generated Videos (AIGV) synthesis.
Models such as Sora~\cite{sora2024}, CogVideoX~\cite{yang2024cogvideox}, and Hunyuan~\cite{kong2025hunyuanvideosystematicframeworklarge} can now generate high-fidelity content with complex dynamics, convincing temporal coherence, and strong adherence to user prompts. 
As these models rapidly evolve, there is an urgent need for standardized evaluation methodologies to ensure comprehensive and fair assessments of their performance. Extending beyond traditional Video Quality Assessment (VQA) metrics, which primarily focuses on perceptual fidelity and technical aspects, the desired AI-Generated Video Evaluation (AIGVE) methods may also consider more perspectives like text-video alignment, motion consistency, temporal coherence, and semantic fidelity. 
Many AIGVE metrics have been proposed, such as TIFA~\cite{hu2023tifa}, VIEScore~\cite{ku2024viescore}, CLIPSim~\cite{hessel2022clipscore}, FVD~\cite{FVD}, FID~\cite{FID}, VideoPhy~\cite{bansal2024videophy}, GSTVQA~\cite{GSTVQA}, PickScore~\cite{PickScore}, Light-VQA+~\cite{Light_VQA_plus}, \etc. each providing distinct assessment perspectives.

Despite this diversity, existing AIGVE approaches face several critical limitations. The field lacks a unified theoretical framework for systematically categorizing evaluation methodologies, hindering a holistic understanding of the evaluation landscape. Research on metric complementarity remains limited, restricting the development of combinations that better align with human perception. Additionally, implementation barriers arise as different metrics require specialized environments and incompatible dependencies, leading to significant engineering overhead. The absence of standardized interfaces forces researchers to repeatedly implement redundant processing steps for each metric, increasing error risks. Further, many existing metrics suffer from poor generalization, as they are often designed for specific datasets, limiting their applicability across different models and domains. These challenges collectively impede systematic benchmark analysis, hinder meta-evaluation research, and prevent fair model comparisons, ultimately slowing the advancement of AIGVE metrics.



To address these challenges, we propose \ours, a unified AIGVE toolkit that provides a structured and extensible evaluation pipeline.
Organized within a novel five-category taxonomy, \ours integrates diverse evaluation methodologies while allowing flexible customization through a modular configuration system.
Inspired by the modular design of widely used toolkits~\cite{Detectron2, MMDetection, MMSegmentation, PyTorch}, \ours operates in a plug-and-play manner, decoupling dataset handling from evaluation logic. This separation allows researchers to configure evaluation workflows via simple configuration files, facilitating the seamless integration of various AIGVE approaches, from video-only assessments to vision-language alignment metrics.
To the best of our knowledge, \ours is the first unified modular architecture specifically designed for AIGVE. By decoupling data loading from metric computation, it enables researchers to effortlessly combine different datasets, models, and evaluation criteria, significantly reducing implementation overhead for new evaluation setups.

At the core of \ours, our \textbf{AIGVELoop} component standardizes the evaluation workflow while maintaining extensibility for custom implementation. The pipeline consists of three key phases: data processing, metric computation, and results aggregation. During data processing, the system extracts and normalizes video content and textual prompts into standardized formats. Our \textbf{Customizable DataLoaders} support various video sources with built-in compatibility for common formats and custom extensions, performing tasks such as resolution normalization, frame sampling, and format conversion. In the metric computation phase, registered \textbf{Modular Metrics} execute through standardized interfaces, efficiently sharing computational resources while supporting both per-video and batch-level assessments. Finally, the results aggregation phase consolidates per-sample outputs into comprehensive reports, facilitating comparative analysis across AIGVE metrics.

Building on this modular framework, \ours introduces a novel five-category taxonomy that systematically classifies existing AIGVE metrics based on their underlying principles. These categories --- Distribution Comparison-Based, Video-Only Neural Network-Based, Vision-Language Similarity-Based, Vision-Language Understanding-Based, and Multi-Faceted Evaluation Metrics --- provide researchers with a structured foundation for selecting complementary metrics that capture different aspects of video quality and alignment. This taxonomy not only clarifies the landscape of existing metrics but also serves as a roadmap for developing new AIGVE methodologies that can be seamlessly integrated into the \ours framework.

To demonstrate the effectiveness of \ours and establish a comprehensive benchmarking resource, we present \textbf{\data}, a large-scale multifaceted benchmark dataset designed to systematically evaluate different video generation models across nine quality dimensions. \data addresses limitations in existing evaluation datasets by incorporating $500$ text prompts, $2,430$ videos generated by five state-of-the-art models (after filtering unrealistic cases), and $21,870$ human evaluation scores spanning nine assessment aspects. Leveraging the structured evaluation capabilities of \ours, this benchmark provides researchers with standardized data for model performance comparison across various conditions and quality dimensions, facilitating more rigorous and reproducible research in AIGVE.


Together, these three components --- the modular \ours architecture, the structured five-category taxonomy, and the multifaceted \data dataset --- form a cohesive ecosystem for AIGVE. \ours provides the technical infrastructure for conducting evaluations, the taxonomy guides the selection and development of appropriate metrics, and \data offers standardized benchmarks for validating these assessments. Our experiments with this integrated approach reveal that even state-of-the-art video generation models struggle with urban scenes and complex interactions while performing better on natural subjects, highlighting key areas for improvement in video generation technology. 
Our contributions are summarized as follows:
\begin{itemize}
    \item We introduce \ours, a unified and modular toolkit standardizing AI-generated video evaluation while enabling flexible, configuration-driven customization.
    \item We propose a five-category taxonomy that systematically classifies AIGVE metrics by their underlying principles, facilitating comprehensive assessment strategies.
    \item We present \data, a multifaceted AIGVE dataset with $500$ prompts, $2,430$ videos, and $21,870$ human-annotated scores across nine evaluation aspects.
    \item We validate the effectiveness of \ours by extensive experiments on diverse video generators, revealing specific strengths and limitations of current models.

\end{itemize}

%% file: content/2-related.tex
\section{Related Work}
\label{sec:related}

\begin{figure*}[!t]
    \centering
    \includegraphics[width=1.0\linewidth]{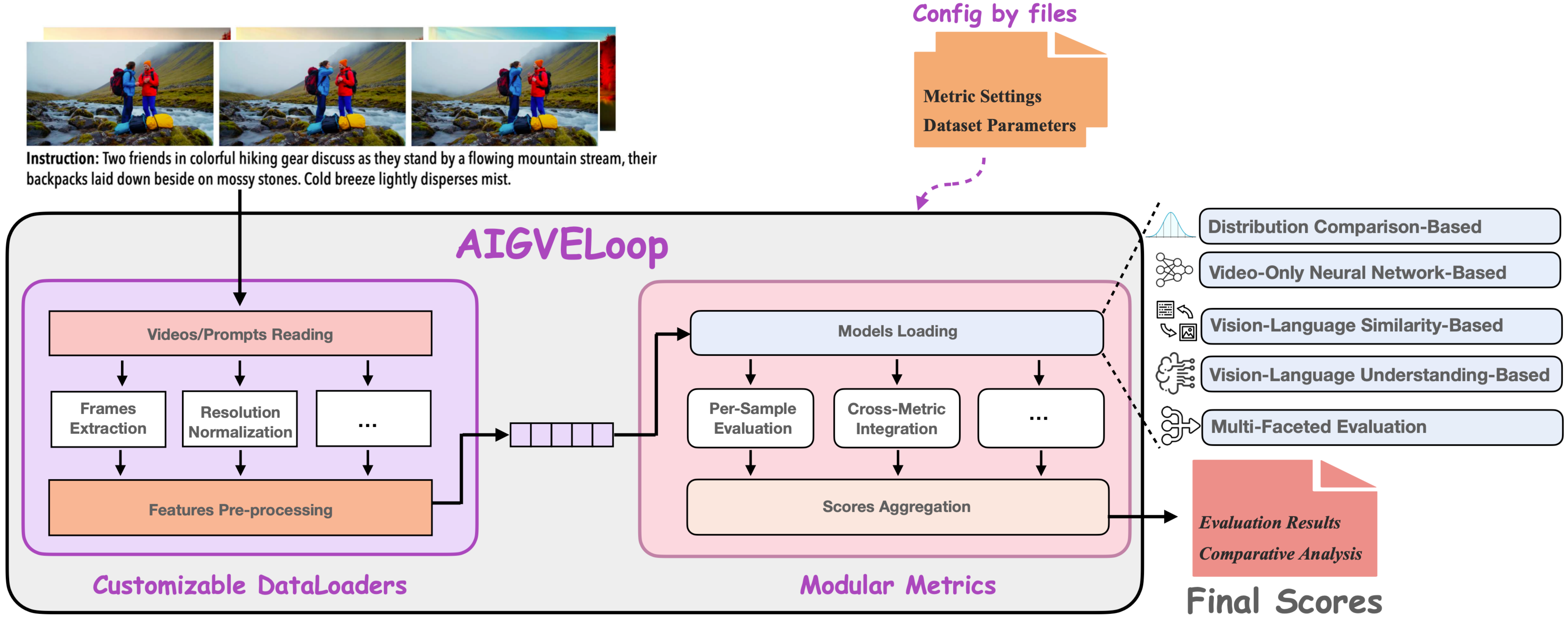}
    \caption{Overall Structure of \ours. Built on MMEngine\cite{mmengine2022}, the framework consists of three core components: 1) Configuration Files that enable customization through Python-based settings for dataset loading, metric selection, and evaluation parameters; 2) Customizable DataLoaders that standardize diverse video formats and extract features while supporting various pre-processing operations; and 3) Modular Metrics organized within our five-category taxonomy. These components interact through the AIGVELoop system which standardizes the workflow with \texttt{process()} and \texttt{compute\_metrics()} interfaces, transforming input data into structured evaluation results. This modular design allows researchers to seamlessly integrate new datasets and metrics without modifying core components.}
    \label{fig:toolkit_structure}
\end{figure*}

\subsection{AI-Generated Video Evaluation (AIGVE)}
Various metrics have been proposed to assess AIGV, covering perceptual quality, realism, and semantic alignment. Traditional Video Quality Assessment methods~\cite{NIQE, Wang2004ImageQA, WANG2004121, GSTVQA, SimpleVQA, Light_VQA_plus} focus on perceptual fidelity, while statistical approaches~\cite{FID, FVD, IS_Score} measure distribution similarity. More recent methods incorporate vision-language models~\cite{hessel2022clipscore, hu2023tifa, li2023blip, PickScore}, temporal consistency analysis~\cite{liu2023evalcrafter, DSG}, and physical realism checks~\cite{bansal2024videophy} to enhance evaluation. 
Additionally, comprehensive methods~\cite{ he2024videoscore, bansal2024videophy, huang_vbench_2023, liu2023evalcrafter} integrate multiple aspects for a more holistic assessment. 
Despite these advancements, existing methods often suffer from dataset dependency and limited flexibility. 
\ours addresses these challenges by providing a unified, extensible evaluation framework to standardize AIGVE.

\subsection{Structured Framework Approaches}
Existing comprehensive AIGVE methods~\cite{huang_vbench_2023, liu2023evalcrafter, he2024videoscore} primarily function as fixed benchmarking protocols rather than adaptable frameworks. 
While they support multi-dimensional evaluation, they suffer from dataset dependency, limited extensibility, and rigid designs, complicating the integration of new AIGVE metrics. 
This forces researchers into custom scripts, increasing overhead and reducing reproducibility. Thus, a modular and extensible AIGVE framework is urgently needed.
Structured, configuration-driven architectures have proven effective in various domains, including 2D object detection\cite{Detectron2, MMDetection}, point cloud perception\cite{MMDetection3D, OpenPCDet}, semantic segmentation~\cite{MMSegmentation}, Pose Estimation~\cite{MMPose}, NLP tasks\cite{HuggingFace, AllenNLP}, speech processing\cite{ESPnet}, robotics simulation\cite{MuJoCo}, and general-purpose ML frameworks~\cite{PyTorch, Keras, Hydra}.
Applying similar principles to the field of AIGVE would facilitate scalable, flexible, and reproducible metrics.



\subsection{Benchmark Datasets in AIGVE}
\label{relate:benchmark}
Various benchmark datasets~\cite{huang_vbench_2023, liu2023evalcrafter, bansal2024videophy, konvid1k, CVD2014, LiveVQC, UGVQ} have been introduced for AIGVE. 
While existing benchmarks have made
notable contributions, they face several critical limitations.
First, most benchmarks~\cite{konvid1k, CVD2014, LiveVQC} prioritize perceptual quality metrics, while few~\cite{liu2023evalcrafter, bansal2024videophy} offer text-video alignment assessment.
Second, existing datasets typically provide aggregate scores rather than detailed, aspect-specific assessments~\cite{liu2023evalcrafter, he2024videoscore, UGVQ}, necessary for comprehensive evaluation. 
Third, these benchmarks frequently include videos from older-generation models that do not reflect state-of-the-art performance~\cite{huang_vbench_2023, liu2023evalcrafter, he2024videoscore}. Additionally, some existing datasets lack coverage of diverse, complex scenarios~\cite{bansal2024videophy, liao2024evaluation}.
These limitations highlight the need for a more comprehensive, structured, and detailed benchmark dataset, motivating our proposed \data dataset.

%% file: content/3-toolkit.tex
\section{Toolkit Architecture}
\label{sec:toolkit}

\subsection{Design Philosophy}
The core design of \ours follows three philosophy: modularity, extensibility, and reproducibility. Modularity decouples data processing from evaluation logic, enabling flexible component integration via configuration files. Extensibility allows seamless incorporation of new metrics, datasets, or techniques without disrupting functionality. Reproducibility is ensured through standardized protocols, data formats, and comprehensive documentation.


\subsection{Modular Framework Design}
To the best of our knowledge, \ours is the first unified framework for AIGVE metrics with a modular and configurable design.
Unlike existing frameworks that impose rigid evaluation pipelines, \ours allows researchers to flexibly integrate diverse metrics and datasets, ensuring adaptability to evolving evaluation needs.
As illustrated in Figure~\ref{fig:toolkit_structure}, \ours standardizes AIGVE while maintaining extensibility.

Further technical details are provided in Supplementary Materials Sec.~\ref{app-usage:dataset}.

\subsubsection{AIGVELoop: The Unified Evaluation Pipeline}
\label{frame:loop}

AIGVELoop is an evaluation pipeline built upon MMEngine's \texttt{BaseLoop} class~\cite{mmengine2022}, standardizing video assessment through three key phases: data processing, metric computation, and results aggregation. It manages the entire evaluation process, from initial data handling to final report generation, ensuring consistency and a structured assessment workflow.
By providing a standardized yet extensible framework, AIGVELoop provides unified interfaces that allow researchers to integrate custom metrics, perform both per-video and batch-level evaluations, and conduct comprehensive comparative analyses across different video generation models.


\subsubsection{Customizable DataLoaders}
\label{frame:dataloader}

\ours introduces a flexible data loading system that supports any dataset formatted using MMFormat JSON annotation file~\cite{mmengine2022}. 
Inherited from PyTorch's \texttt{Dataset} class~\cite{PyTorch}, our Customizable DataLoaders offer a robust framework for video data management, handling diverse video sources reading, prompt retrieval, resolution normalization, frame sampling and padding, format conversion, feature extraction and pre-processing, \etc.
The system processes video files by reading JSON annotation files, extracting frames via OpenCV~\cite{opencv}, and converting into standardized Torch Tensor~\cite{PyTorch} representations, ensuring consistent input across different evaluation scenarios.
The Customizable DataLoaders simplify dataset integration.
Researchers only need to implement a lightweight \texttt{\_\_getitem\_\_()} method, while the framework automates batch processing, feature extraction, and metric evaluation. 
By decoupling data loading from evaluation logic, \ours minimizes implementation overhead, facilitates custom dataset integration, and allows researchers to focus on metric development rather than data preparation complexities.



\subsubsection{Modular Metrics}
\label{frame:metrics}

\ours implements all AIGVE metrics as standalone, extensible modules built upon MMEngine's \texttt{BaseMetric} class~\cite{mmengine2022}.
Modular Metrics support various model-related operations such as model loading, dynamic feature extraction, per-sample evaluation, score aggregation, flexible computational resource management, \etc.
Each metric just need to follow standardized interfaces to customize its operations, such as consistent \texttt{process()} for per-sample evaluation and \texttt{compute\_metrics()} for scores aggregation.
Modular Metrics also enables cross-metric integration, which is important for robust evaluation. 

Through extensible interfaces, researchers can easily incorporate diverse evaluation methods with minimal modifications of the underlying architecture, while benefiting from resource-efficient computation. 
The decoupled Modular Metrics architecture separates AIGVE logic from infrastructure concerns, enabling seamless integration into AIGVELoop.
Cross-metric integration facilitates interactions between different AIGVE metrics, allowing complementary metrics to work together for more comprehensive assessments.
The modular design lowers barriers to develop new AIGVE methodologies while ensuring consistency, ultimately advancing AIGVE research.






\subsubsection{Configuration-Driven Execution}
\label{frame:config}

\ours employs Python-based configuration files to manage execution settings, allowing users to define runtime parameters for their dataloader, evaluation metric, and any other components within AIGVELoop and MMEngine runner. 
Various video-specific parameters can therefore be easily set, such as data source, data sampling method, frame sampling rate, feature extraction method, pre-trained model path, metric integration strategy, input modalities, customized weights, \etc. 
Configuration files are hierarchically structured, eliminating scattered modifications across multiple files and streamlining parameter management.

This centralized, configuration-driven approach allows researchers to manage the entire evaluation process from a single, well-organized file.
Settings can be inherited and reused, reducing redundancy while enabling flexible adjustments for specific experiments.
Researchers can seamlessly switch between datasets and metrics without modifying core code, ensuring consistency, improving reproducibility, and facilitating systematic comparisons.


\subsection{Unified Categorization of AIGVE Metrics}
AIGVE's key challenge is its multi-faceted nature, requiring evaluation across diverse dimensions from visual quality to semantic alignment.
\ours addresses this by proposing a novel five-category taxonomy that systematically organizes existing AIGVE metrics based on their underlying principles. 
This structured categorization provides a comprehensive foundation for benchmarking AIGV, enabling researchers to analyze video quality through complementary perspectives rather than isolated metrics. 
These five categories are detailed in Sec~\ref{sub:dis-based} - Sec~\ref{sub:multi-faceted}, with all supported metrics listed in Supplementary Materials Sec.~\ref{app:supported}.

\subsubsection{Distribution Comparison-Based Metrics} 
\label{sub:dis-based}
Distribution comparison metrics evaluate the statistical similarity between generated and real video distributions in feature space. \ours incorporates key metrics such as FID~\cite{FID}, FVD~\cite{FVD}, Inception Score~\cite{IS_Score}, \etc. These metrics provide a macro-level assessment of generation quality. 

\subsubsection{Video-Only Neural Network-Based Metrics}
\label{sub:nn-based}
Video-only neural metrics assess perceptual quality without considering textual prompts. These metrics leverage deep neural networks trained on human perception data to evaluate technical aspects such as spatio-temporal consistency, motion smoothness, structural integrity, sharpness, and artifact presence~\cite{liu2024surveyaigve}.
\ours integrates existing metrics such as GSTVQA~\cite{GSTVQA}, Simple-VQA~\cite{SimpleVQA}, Light-VQA+~\cite{Light_VQA_plus}, \etc. These metrics identify technical deficiencies, including motion artifacts and perceptual coherence in generated videos, making them well-suited for evaluating AIGV fidelity. 

\subsubsection{Vision-Language Similarity-Based Metrics}
\label{sub:vl-sim-based}
Text-video alignment is crucial for evaluating prompt-conditional generated videos. 
These vision-language similarity metrics compute semantic similarity between generated videos and their textual descriptions using embedding-based calculations. 
\ours incorporates foundational cross-modal metrics including CLIPSim~\cite{hessel2022clipscore},  BLIPSim~\cite{li2023blip}, PickScore~\cite{PickScore}, \etc. These metrics signal how well AIGV matches high-level concepts in prompts.

\subsubsection{Vision-Language Understanding-Based Metrics}
\label{sub:vl-understanding-based}
Beyond feature similarity, vision-language understanding metrics employ more complex reasoning techniques to evaluate semantic accuracy and adherence to prompt details.
\ours supports understanding-based metrics including VIEScore~\cite{ku2024viescore}, TIFA~\cite{hu2023tifa}, DSG~\cite{DSG}, \etc. 
These metrics go beyond simple embedding comparisons to detect nuanced inconsistencies in object interactions, motion accuracy, and scene plausibility. 
Some methods~\cite{DSG} further introduce scene graphs to model spatial and relational attributes within videos, enabling compositional verification, making them valuable for AIGV assessment.

\subsubsection{Multi-Faceted Evaluation Metrics}
\label{sub:multi-faceted}
Multi-faceted metrics evaluate multiple quality dimensions simultaneously. These metrics provide holistic evaluations of AIGV aligned with human judgment by considering diverse aspects of video quality.
\ours incorporates various evaluation frameworks such as VideoPhy~\cite{bansal2024videophy}, VideoScore~\cite{he2024videoscore}, \etc. 
Combining spatial, temporal, and linguistic evaluation in a unified framework, these metrics bridge the gap among different evaluation dimensions, offering comprehensive AIGV assessments.\newline

Organized into this five-category taxonomy, \ours provides the first systematic, configurable, and extensible solution for AIGVE, addressing the fragmentation in existing evaluation methodologies.
Its modular design enables researchers to assess trade-offs across quality dimensions, identifying model strengths and weaknesses.
We hope \ours organized in this taxonomy will lay a foundation for future research, fostering a deeper understanding of their capabilities and guiding advancements in next-generation video models.







%% file: content/4-benchmark.tex
\section{\data Benchmark Dataset}
\label{sec:benchmark}


To address those limitations mentioned in Sec.~\ref{relate:benchmark}, we present \textbf{\data}, a multifaceted benchmark dataset that systematically tackles these challenges. 
Building on \ours's modular framework and our five-category taxonomy, \data provides standardized data for thorough, multi-dimensional assessment.
While \ours offers the technical infrastructure for implementing diverse evaluation methodologies, \data provides the carefully chosen content necessary to apply these methods consistently. 

Drawing from our structured evaluation framework, the benchmark decomposes evaluation into two fundamental dimensions: alignment with human perception and alignment with human instructions. These dimensions are further refined into $9$ distinct, quantifiable aspects that for precise assessment. Through rigorous human annotation, our benchmark dataset comprising $500$ diverse prompts, $2,430$ videos generated by five state-of-the-art video generation models, and $21,870$ fine-grained assessment scores.

\subsection{Dataset Construction}
Following the classic AIGVE benchmark dataset collection pipeline~\cite{liu2024surveyaigve}, our benchmark dataset construction split into 4 phases: 1) Instruction Generation, 2) Instruction Validation, 3) Video Generation, and 4) Video Evaluation. Below we introduce each phase in detail.

\begin{table}[ht]
    \centering
    \small 
    \renewcommand{\arraystretch}{1.0} 
    \setlength{\tabcolsep}{4pt} 
    \scalebox{0.75}{
    \begin{tabular}{l|c c}
        \toprule
        \textbf{Category} & \textbf{Objects} & \textbf{Dynamic} \\
        \midrule
        \textbf{\textit{Global View}} & 
        \begin{tabular}[c]{@{}l@{}}
            Mountains, Oceans, Plains, \\ 
            Rivers, Lakes, Deserts, Cities, \\ 
            Architecture, Streets 
        \end{tabular} & 
        \begin{tabular}[c]{@{}l@{}}
            Camera Movements, Daylight \\ 
            Transitions, Weather Changes, \\ 
            Seasonal Shifts 
        \end{tabular} \\
        \midrule
        \textbf{\textit{Close Shot}} & 
        \begin{tabular}[c]{@{}l@{}}
            Humans, Animals, Plants, Objects 
        \end{tabular} & 
        \begin{tabular}[c]{@{}l@{}}
            Single-Object Actions, \\ 
            Multiple-Object Interactions 
        \end{tabular} \\
        \bottomrule
    \end{tabular}    
    }
    \caption{Summary of Instruction Categories}
    \label{tab:instructions_description}
\end{table}

\subsubsection{Instruction Generation and Validation}
Since the primary objective of \data is to assess AI-generated videos against human perception and instructions, we need structured, quantifiable instructions with detailed elements and interactions. Unlike existing text-to-video datasets \cite{Bain21, xu2016msr} that either focus too narrowly or too broadly, we implement a categorize-then-generate pipeline that constrains objects and interactions within predefined scopes. 
This approach leverages Large Language Models with carefully crafted prompts to produce diverse yet structured instructions covering real-world scenes, enabling more controlled and rigorous evaluation.

Specifically, our instructions are divided into two categories: \textbf{\textit{global view}} and \textbf{\textit{close shot}}. The \textbf{\textit{global view}} instructions describe panoramic, B-roll-like scenes commonly found in landscape photography. These instructions evaluate a video generation model's understanding of environmental contexts and its ability to capture natural interactions, such as waves crashing onto the shore, leaves falling with the season change, and shadows elongating at sunset.

The primary objects in these instructions include natural and urban environments, including mountains, oceans, plains, rivers, lakes, deserts, cities, architectural structures, and streets. To avoid static videos, we ensure each instruction includes at least one dynamic element. These dynamic elements include camera movements, daylight transitions, weather changes, and seasonal shifts. A detailed description of each dynamic component, along with exemplar prompts, can be found in the Supplementary Material Sec.~\ref{app:dynamic_des}.

Similarly, the \textbf{\textit{close shot}} instructions mainly evaluate a video generation model's understanding of individual objects and their interactions. The close-shot instructions could include objects within human, animal, plant, and object. We categorize their dynamic into single-object actions and multiple-object interactions. Tab. \ref{tab:instructions_description} summarizes the object and dynamic options for each instruction category.

Using predefined objects and dynamics, we carefully designed a prompt to instruct GPT-4o \cite{openai2024gpt4technicalreport} to generate diverse instructions while labeling the included objects and dynamics. The prompts we designed to generate the instructions. are presented in Supplementary Material Sec.~\ref{app:instruction_prompt}. To ensure the quality and accuracy of the generated instructions, we conducted a rigorous human evaluation process to verify that the instructions were unique and that the objects and dynamics were correctly labeled. As a result, we collect 500 distinct video generation instructions.

\begin{table}[!t]
    \centering
    \small
    \begin{tabular}{l|r r r}
        \toprule
        \textbf{Model} & \textbf{Resolution} & \textbf{Frame Rate} & \textbf{Duration} \\
        \midrule
        \textbf{CogVideoX} \cite{yang2024cogvideox} & 1360 $\times$ 768 & 16 fps & 5s \\
        \textbf{Genmo} \cite{genmo2024mochi} & 848 $\times$ 480 & 30 fps & 5s \\
        \textbf{Hunyuan} \cite{kong2025hunyuanvideosystematicframeworklarge} & 960 $\times$ 544 & 24 fps & 5s \\
        \textbf{Pyramid} \cite{jin2024pyramidal} & 1280 $\times$ 768 & 24 fps & 5s \\
        \textbf{Sora} \cite{sora2024} & 854 $\times$ 480 & 30 fps & 5s \\
        \bottomrule
    \end{tabular}
    \caption{Specifications of Generated Videos from each Model.}
     \label{tab:video_specs}
\end{table}

\subsubsection{Video Generation}
To ensure a comprehensive and objective evaluation reflecting current state-of-the-art in video generation research, we selected 5 advanced models from both open-source and commercial categories: Pyramid \cite{jin2024pyramidal}, CogVideoX \cite{yang2024cogvideox}, Genmo \cite{genmo2024mochi}, Hunyuan \cite{kong2025hunyuanvideosystematicframeworklarge}, and Sora \cite{sora2024}. Unlike prior work that standardizes technical specifications, such as resolution and frame rate, we align with human preferences for these specifications in our \data benchmark dataset. To achieve this, we only set the video duration to a uniform 5 seconds to ensure that models generate videos of sufficient length, enabling us to test long-range temporal consistency. For other parameters, we implement the highest possible configurations supported by our computational resources, financial budget, and model capabilities to evaluate a diverse set of videos in the next phase. The detailed specifications for each model are presented in Tab.~\ref{tab:video_specs}.

\subsubsection{Video Evaluation}
Building on the advancements in AIGVE research, we adopt the metric decomposition schema proposed by VBench \cite{huang_vbench_2023} and EvalCrafter \cite{liu2023evalcrafter}. We systematically break down the evaluation of alignment with human perception and instructions into 9 multifaceted and easy-to-quantify aspects: 1) Technical Quality, 2) Dynamic, 3) Consistency, 4) Physics, 5) Element Presence, 6) Element Quality, 7) Action/Interaction Presence, 8) Action/Interaction Quality, and 9) Overall. Detailed explanations of each aspect are provided in Supplementary Material Tab. \ref{tab:metrics}.

We engaged 4 expert graduate student annotators for the scoring process. To ensure consistency and robust scoring, we conducted an online calibration session with systematic training, detailed explanations of scoring aspects, and multiple rounds of practice scoring with debriefings. Annotators then evaluated a 50-sample dataset to calculate the Inter-Annotator Agreement, achieving a Krippendorff's $\alpha$ of 0.7035, indicating well-calibrated and reliable scores.

\subsection{Dataset Statistics}

\begin{figure*}[!t]
    \centering
    \includegraphics[width=0.88\linewidth]{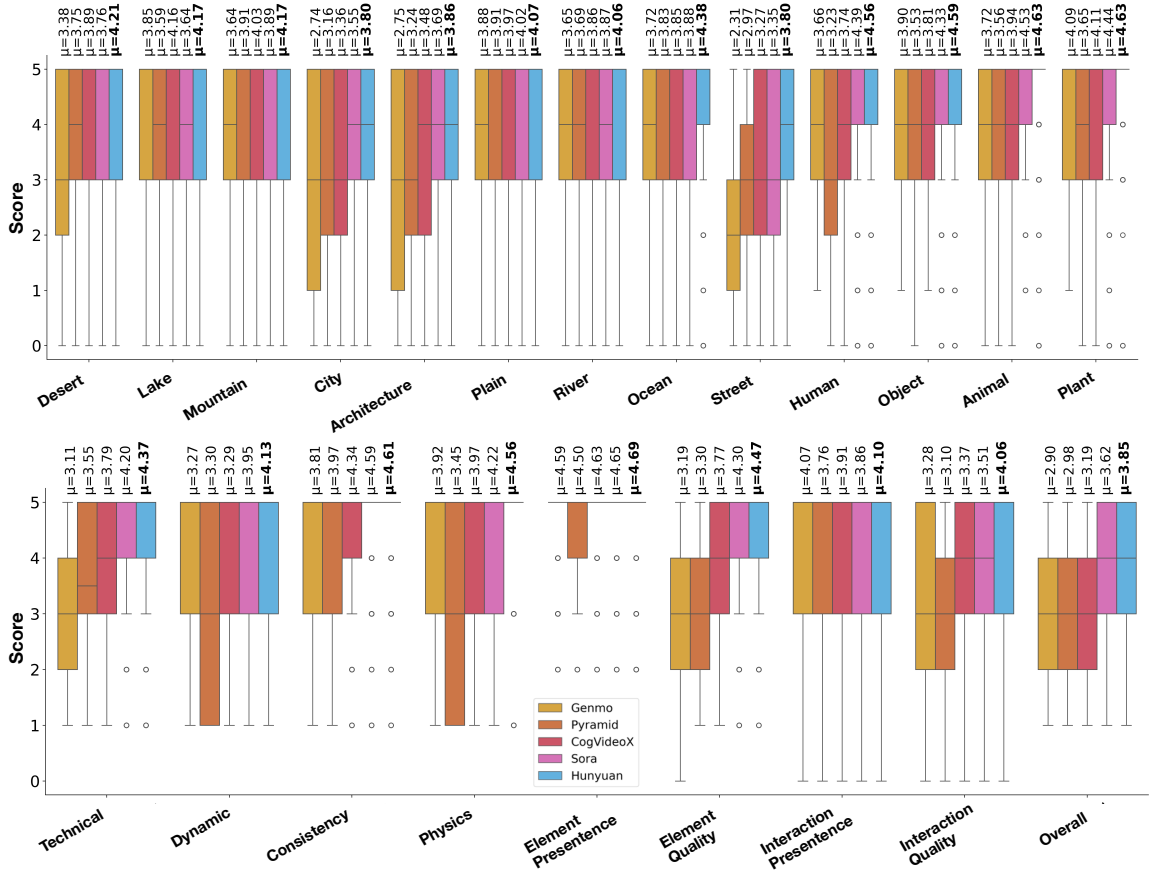}
    \caption{Score distribution across different models. The top plot illustrates the models' overall performance across various object categories, while the bottom plot presents their performance across different evaluation metrics. $\mu$ represents the mean score.}
    \label{fig:score_dist}
\end{figure*}

After evaluating videos, we filtered out videos that were significantly unrealistic, such as 3D/2D animations, and those with severe distortions hindering evaluation. Ultimately, our \data benchmark comprises $500$ curated prompts and $2,430$ videos, totaling $21,870$ individual scores. The creation of the dataset required approximately $1,000$ hours to generate and score videos. The prompts are evenly divided into $250$ global view prompts and $250$ close shot prompts, with the subject and dynamic distributions illustrated in Supplementary Material Fig.~\ref{fig:subject_dist}.

%% file: content/5-experiment.tex
\section{Experiment}
\label{sec:exp}

\subsection{Model Performance Analysis}
With multifaceted human evaluation scores, we analysis state-of-the-art video generation models. Fig.~\ref{fig:score_dist} shows performance distributions of our five selected models across object categories and evaluation aspects.
By comparing individual model performance, our analysis reveals that the latest commercial model, Sora, and the open-sourced Hunyuan model significantly outperform earlier models. Notably, the open-sourced \textbf{Hunyuan model achieves the highest mean overall scores across all subjects and evaluation aspects}. It excels in maintaining consistency and adhering to real-world physics while ensuring objects mentioned in the instructions are presented with high quality. This highlights its strong capability to generate videos that align well with both human perception and the instructions.

\begin{figure}[!t]
    \centering
    \includegraphics[width=1.0\linewidth]{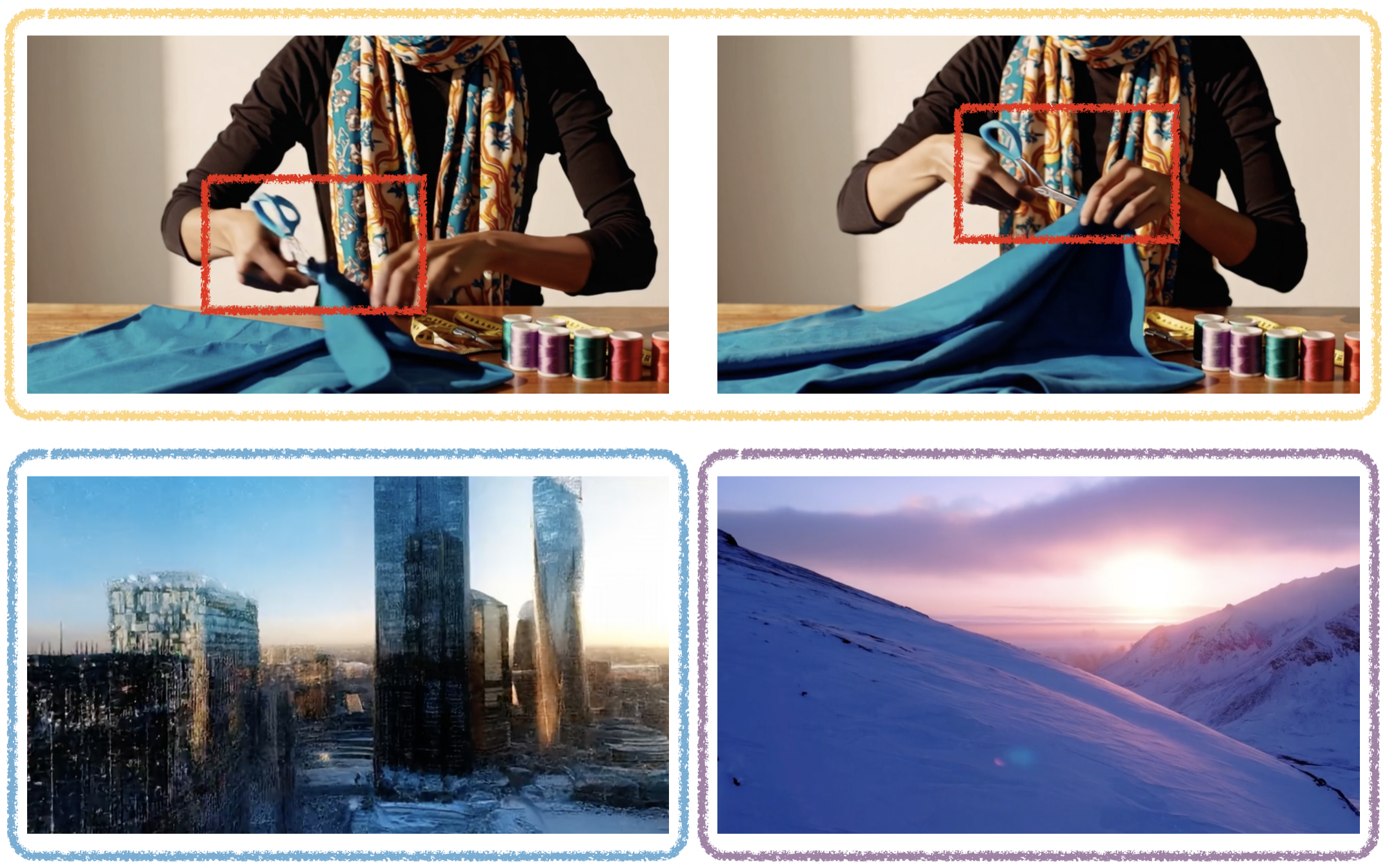}
    \caption{Case Study of AIGVE-Bench. The top row shows that state-of-the-art video generation models excel at high-fidelity elements but struggle with interaction quality (red squares). The bottom row highlights the models are better on generating natural scenes rather than urban views. }
    \label{fig:case_study_paper}
\end{figure}

By examining aggregated model performance, we observe inherent trade-offs in their current implementations. Aggregating performance across aspects reveals that existing models are primarily optimized to ensure the presence of elements mentioned in the instructions and maintain object consistency throughout the video. However, their ability to generate high-quality elements varies significantly. In contrast, \textbf{interaction quality emerges as the weakest-performing category}, as shown in the top row in Fig. \ref{fig:case_study_paper}. 

We attribute this to the dominant diffusion-based architecture \cite{ho2020denoising}, which typically incorporates a CLIP model \cite{radford2021learningtransferablevisualmodels} as the multimodal encoder and a specialized attention mechanism as the diffusion backbone \cite{xing2024survey}. While CLIP effectively aligns textual and visual modalities, ensuring the inclusion of objects, and the attention helps maintain object consistency, less research has focus on enhancing the quality of generated elements and interactions. Our analysis underscores the need for future advancements in implementing such components into the diffusion architecture.

By analyzing model performance across subject categories, we conclude that \textbf{state-of-the-art models are generally proficient at generating natural subjects} and environments, such as plants, animals, and plains. \textbf{However, their performance decreases significantly when generating urban scenes}, such as cities, architecture, and streets, as compared in the lower row in Fig.~\ref{fig:case_study_paper}. These scenes require the accurate inclusion of intricate details, such as building textures, reflections on glass surfaces, and complex interactions involving large amounts of pedestrians and vehicles.

Beyond the architectural limitations discussed above, 
we hypothesize that this performance gap is also influenced by training data distribution. Training datasets likely contain abundance natural landscapes while underrepresenting structured urban environments, leading to weaker generalization to these more complex, detail-rich scenarios. More case studies are in Supplementary Material Sec.~\ref{app:case_study}.

\subsection{Exemplar Practice: Evaluating Video Generation Models with \textbf{\ours}}

Our comprehensively integrated toolkit enables the efficient evaluation of video generation models using widely adopted automatic metrics. In this section, we demonstrate the practical application of \ours. 

We begin by computing a comprehensive set of metrics for each instruction-video pair in the test split and report the averaged results in Tab.~\ref{tab:auto_scores}, providing robust baselines for future research. \ours offers simple hyperparameter configuration and simultaneous submission of multiple metrics, thereby significantly enhancing evaluation efficiency.

Despite rapid advancements in video generation, current metrics remain insufficient for comprehensively evaluating AIGV across multiple aspects \cite{liu2024surveyaigve}. However, given the necessity of assessing current video generation models, a practical approach during this transition period is to leverage existing metrics to evaluate distinct aspects of generated videos. To examine the effectiveness of these metrics, we conducted experiments to determine the most correlated automatic metrics for each evaluation aspect proposed in our \data benchmark dataset.

\input{tables/autometric_results}

\input{tables/recommened_metrics}

Specifically, we compute Spearman’s Rank Correlation Coefficient (SRCC) between the automatic metrics listed in Tab.~\ref{tab:auto_scores} and each evaluation aspect in \data dataset. For each aspect, we identify the metric with the highest correlation as the recommended metric to estimate it. The results are summarized in the middle columns of Tab.~\ref{tab:recom_metrics}.

Our analysis reveals that most existing automatic metrics show weak positive correlations with human judgments, potentially leading to inaccurate evaluations. To address this limitation, we introduce a straightforward yet effective regression-based approach that improves correlation with human scores. Although developing a more robust modeling method is a goal for future work, our current approach marks a significant step toward more reliable evaluation.

To achieve this, we use the most correlated metric for each aspect as features and apply linear regression to predict human scores for each evaluation category. Results in the $\text{SRCC}_{reg}$ column of Tab.~\ref{tab:recom_metrics} demonstrate that incorporating information from multiple metrics significantly enhances the correlation with human judgments across all aspects. Consequently, using \ours not only provides efficient computation of individual metric scores but also delivers a more reliable overall evaluation by leveraging the complementary strengths of existing metrics.

%% file: tables/autometric_results.tex
\begingroup
\setlength{\tabcolsep}{4pt}
\begin{table}[!t]
\centering
\small
\scalebox{0.78}{
\begin{tabular}{l|ccccc}
\toprule
\textbf{Metric [range]}        &  \textbf{Genmo} &   \textbf{Pyramid} & \textbf{CogVideoX} &     \textbf{Sora} &   \textbf{Hunyuan} \\
\midrule
SimpleVQA\cite{SimpleVQA} [0, 5] &  3.2060 &   \textbf{3.8201} &   3.5791 &  3.7758 &  3.7092 \\
GSTVQA\cite{GSTVQA} [0, 1] &  0.2061 &   0.2236 &   0.2173 &  \textbf{0.2679} &  0.2175 \\
LightVQA+\cite{Light_VQA_plus} [0, 100] & 58.407 & 64.273 & \textbf{73.541} & 64.693 & 64.396 \\
\midrule
CLIPScore\cite{hessel2022clipscore} [0, 1] &  0.3259 &   0.3293 &   0.3305 &  0.3263 &  \textbf{0.3313} \\
DSGScore\cite{DSG} [0, 1]  &  \textbf{0.5251} &  0.3941 &   0.4138 &  0.4511 &  0.4646 \\
PickScore\cite{PickScore} [0, 1]  &  0.2084 &   0.2105 &   0.2091 &  0.2114 &  \textbf{0.2123} \\
VIEScore\cite{ku2024viescore} [0, 10] &  4.5573 &  4.7395 &   4.7251 &  \textbf{5.3102} &  5.1532 \\
CLIPTemp\cite{liu2023evalcrafter} [0,1]   &  0.9521 &   0.9511 &   \textbf{0.9571} &  0.9457 &  0.9535 \\
TIFA\cite{hu2023tifa} [0, 1]      &  0.6757 &   0.6947 &   \textbf{0.7222} &  0.7025 &  0.6995 \\
\midrule
VideoPhy\cite{bansal2024videophy} [0, 1]  &  0.1601 &  0.1644 &   0.1660 &  0.1547 &  \textbf{0.1799} \\
VideoScore\cite{he2024videoscore} [0, 4] &  \textbf{3.5492} &   2.7216 &   3.1028 &  2.9376 &  2.8064 \\
\bottomrule
\end{tabular}
}
\caption{Evaluation results for state-of-the-art video generation models using the automatic metrics integrated into our toolkit. The values in [$\cdot$] indicate the range of each metric.}
\label{tab:auto_scores}
\end{table}
\endgroup

%% file: tables/recommened_metrics.tex
\begin{table}[t!]
    \centering
    \small
    \scalebox{0.85}{

      \begin{tabular}{l|c|lr|c}
        \toprule
        \textbf{Aspect} & \textbf{$\text{SRCC}_{rand}$} & \textbf{Metric} & \textbf{SRCC} & \textbf{$\text{SRCC}_{Reg}$} \\
        \midrule
        Technical Quality & -0.0671 & SimpleVQA & 0.2449 & \textbf{0.3350 }\\
        Dynamic & 0.0428 & GSTVQA & 0.1340 & \textbf{0.1614} \\
        Consistency & -0.0355 & CLIPTemp & 0.2768 & \textbf{0.3309} \\
        Physics & -0.0162 & CLIPTemp & 0.2351 & \textbf{0.2409} \\
        Element Presence & -0.0303 & PickScore & 0.2624 & \textbf{0.2878} \\
        Element Quality & 0.0142 & CLIPTemp & 0.2512 & \textbf{0.3292} \\
        Action Presence & -0.0516 & CLIPScore & 0.2384 & \textbf{0.2608} \\
        Action Quality & 0.0144 & CLIPTemp & 0.2019 & \textbf{0.2127} \\
        Overall & -0.0319 & PickScore & 0.2428 & \textbf{0.3561} \\
        \bottomrule
    \end{tabular}

    }
  
    \caption{The recommended metric for each evaluation aspect in AIGVE-Bench. $\text{SRCC}_{rand}$ represents the SRCC between human label and random scores. $\text{SRCC}_{reg}$ represents the SRCC between human label and the regression method scores.}
    \label{tab:recom_metrics}
\end{table}

%% file: content/6-conclusion.tex
\section{Conclusion}
\label{sec:conclusion}

We introduce \ours, a unified, extensible framework for AIGVE, addressing prior limitations through a modular, taxonomy-driven architecture. 
By combining multiple evaluation paradigms and benchmarking capabilities, \ours standardizes AIGVE while enabling flexible customization. 
Our \data dataset further enhances the toolkit by providing a large-scale, human-annotated benchmark dataset. Extensive experiments show \ours enables robust, reliable, and scalable video evaluation, revealing current models excel with natural subjects but struggle with urban scenes and complex interactions, thus advancing the field of AIGVE research.

%% file: content/Appendix.tex
\clearpage
\onecolumn
\setcounter{page}{1}
\setcounter{section}{0}
\setcounter{table}{4}
\setcounter{figure}{3}

\begin{center}
    \Large
    \textbf{\thetitle}\\
    \vspace{0.5em}Supplementary Material \\
    \vspace{1.0em}
\end{center}


\section{Instruction Generation Prompt}
\label{app:instruction_prompt}
\input{tables/prompt_generation}

\section{Detailed Description of Dynamics}
\label{app:dynamic_des}

When constructing the global view instructions in AIGVE-Bench, we ensure that each instruction incorporates at least one dynamic element. These dynamic elements enrich the generated videos by adding realistic motion and environmental variability.
\begin{itemize}
    \item \textbf{Camera Movements:} 
    Dynamic camera movements simulate realistic cinematography by incorporating actions such as panning, tilting, tracking, and zooming. These movements can follow a moving subject, explore the environment, or provide dramatic close-ups to enhance visual storytelling. \\
    \emph{\textbf{Exemplar Instruction:}} "The vastness of the ocean reflects golden hues during dusk while gentle waves ripple beneath. Marine birds soar above as the camera follows their graceful flight." \\ 

    \item \textbf{Daylight Transitions:} 
    Daylight transitions capture the natural evolution of lighting conditions, emulating changes from dawn to dusk or the gradual shift in light during sunrise or sunset. These transitions add a temporal dimension to the scene, highlighting the interplay of natural light. \\
    \emph{\textbf{Exemplar Instruction:}} "A grandiose bridge spans across a bustling urban waterway, with boats and small ferries navigating actively. The view transitions from afternoon, highlighting daily commuters, to evening, against a backdrop of twinkling city lights." \\

    \item \textbf{Weather Changes:} 
    Incorporating weather changes introduces variability and atmosphere into the scene. This dynamic element includes conditions such as rain, snow, fog, and clear skies, each affecting the mood and visual composition differently. \\
    \emph{\textbf{Exemplar Instruction:}} "A Delicate Monsoon spectacle at a lush river's edge, as distant thunderclouds release a majestic downpour, polka-dotted shadows racing across the deep-current waters abound with life." \\

    \item \textbf{Seasonal Shifts:} 
    Seasonal shifts reflect the natural progression of the year, influencing the color palette and overall mood of the scene. Transitions among spring, summer, autumn, and winter provide contextual depth and visual diversity. \\
    \emph{\textbf{Exemplar Instruction:}} "A sprawling plain stretches into the distance with fields of flowers swaying in the breeze. Birds fly above as the scene changes from the vibrant colors of spring to the golden tones of autumn." \\
\end{itemize}

\newpage
\section{AIGVE-Bench Subject and Dynamic Distribution}

\begin{figure}[H]
    \centering
    \includegraphics[width=0.90\linewidth]{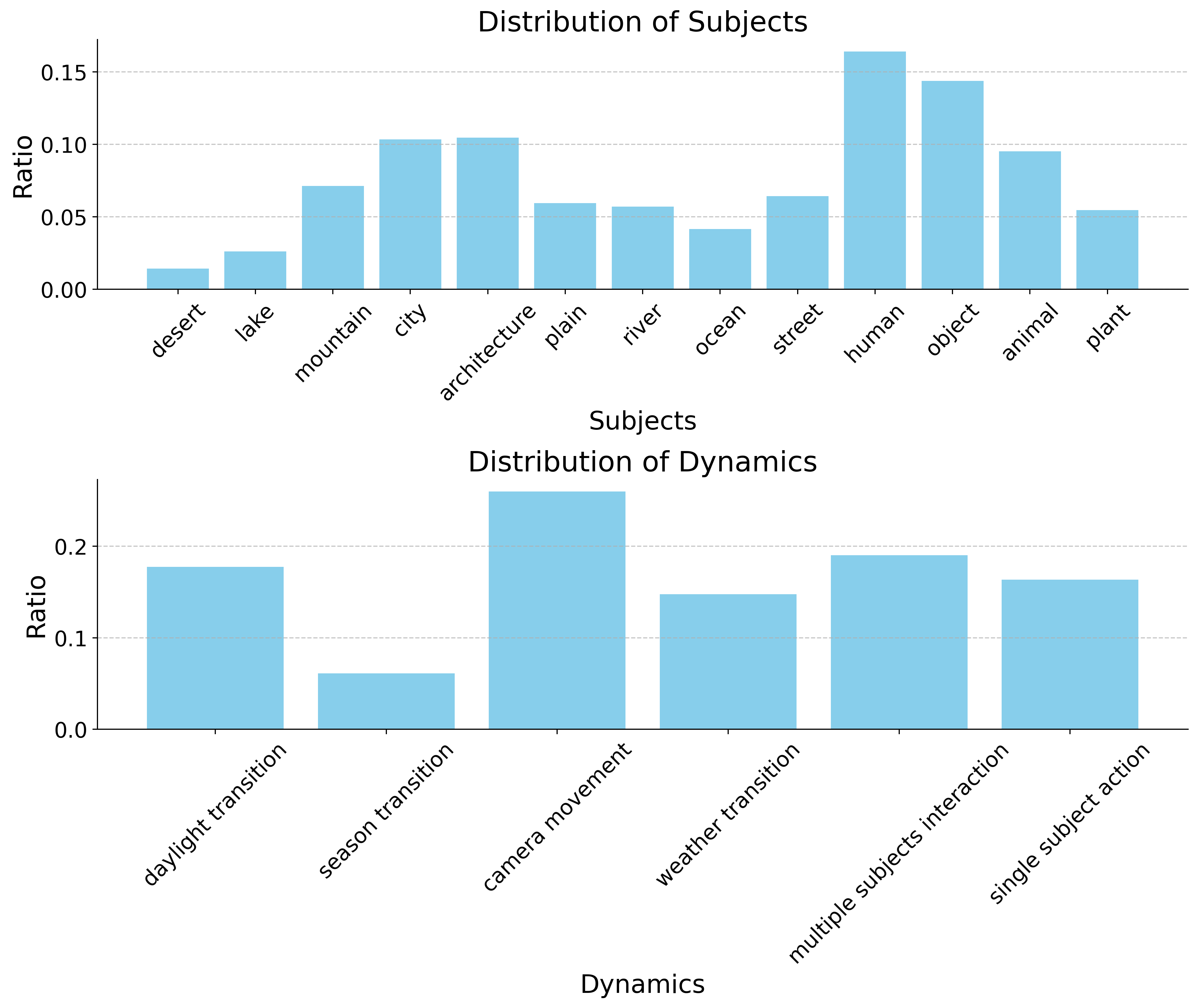}
    \caption{Subject and Dynamic Distribution of AIGVE-Bench Benchmark Dataset.}
    \label{fig:subject_dist}
\end{figure}

\newpage
\section{Detailed Description of AIGVE-Bench Evaluation Aspects}

\input{tables/metrics}

\newpage
\section{Case Study}
\label{app:case_study}
\begin{figure*}[!h]
    \centering
    \includegraphics[width=0.70\linewidth]{figures/case_study.png}
    \caption{Case Study of Human Evaluation for Video Generation Models. TQ: Technical Quality, DYM: Dynamics, CONS: Consistency, PHY: Physics, EP: Element Presence, EQ: Element Quality, AP: Action Presence, AQ: Action Quality, OR: Overall.}
    \label{fig:case_study}
\end{figure*}

\section{\ours Supported Frameworks}
\label{app:supported}

\begin{table}[h!]
    \centering
    \small
    \renewcommand{\arraystretch}{1.2}
    \scalebox{0.95}{
    \begin{tabular}{>{\centering\arraybackslash}m{4.5cm}|p{12cm}}
        \toprule
        \textbf{Category} & \textbf{Metrics and Descriptions} \\
        \midrule
        \multirow{2}{=}{\centering \textbf{Distribution Comparison-} \\ \textbf{Based Metrics}} & 
        \begin{itemize}[leftmargin=0.4cm]
            \item \textbf{Frechet Inception Distance (FID) \cite{FID}:} Quantifies the similarity between real and generated video feature distributions by measuring the Wasserstein-2 distance.
            \item \textbf{Frechet Video Distance (FVD) \cite{FVD}:} Extends the FID approach to video domain by leveraging spatio-temporal features extracted from action recognition networks.
            \item \textbf{Inception Score (IS) \cite{IS_Score}:} Evaluates both the quality and diversity of generated content by analyzing conditional label distributions.
        \end{itemize} \\
        \rowcolor[HTML]{F2F2F2}
        \multirow{2}{=}{\centering \textbf{Video-only Neural} \\ \textbf{Network-Based Metrics}} & 
        \begin{itemize}[leftmargin=0.4cm]
            \item \textbf{Generalized Spatio-Temporal VQA (GSTVQA) \cite{GSTVQA}:} Employs graph-based spatio-temporal analysis to assess video quality.
            \item \textbf{Simple Video Quality Assessment (Simple-VQA) \cite{SimpleVQA}:} Utilizes deep learning features for no-reference video quality assessment.
            \item \textbf{Light Video Quality Assessment Plus (Light-VQA+) \cite{Light_VQA_plus}:} Incorporates exposure quality guidance to evaluate video quality.
        \end{itemize} \\
        \multirow{2}{=}{\centering \textbf{Vision-Language} \\ \textbf{Similarity-Based Metrics}} & 
        \begin{itemize}[leftmargin=0.4cm]
            \item \textbf{CLIP Similarity (CLIPSim) \cite{liu2023evalcrafter}:} Leverages CLIP embeddings to measure semantic similarity between videos and text.
            \item \textbf{CLIP Temporal (CLIPTemp) \cite{liu2023evalcrafter}:} Extends CLIPSim by incorporating temporal consistency assessment.
            \item \textbf{Bootstrapped Language-Image Pre-training Similarity (BLIPSim) \cite{li2023blip}:} Uses advanced pre-training techniques to improve video-text alignment evaluation.
            \item \textbf{PickScore \cite{PickScore}:} Incorporates human preference data to provide more perceptually aligned measurement of video-text matching.
        \end{itemize} \\
        \rowcolor[HTML]{F2F2F2}
        \multirow{2}{=}{\centering \textbf{Vision-Language} \\ \textbf{Understanding-Based Metrics}} & 
        \begin{itemize}[leftmargin=0.4cm]
            \item \textbf{Video Information Evaluation Score (VIEScore) \cite{ku2024viescore}:} Provides explainable assessments of conditional image synthesis.
            \item \textbf{Text-Image Faithfulness Assessment (TIFA) \cite{hu2023tifa}:} Employs question-answering techniques to evaluate text-to-image alignment.
            \item \textbf{Davidsonian Scene Graph (DSG) \cite{DSG}:} Improves fine-grained evaluation reliability through advanced scene graph representations.
        \end{itemize} \\
        \multirow{2}{=}{\centering \textbf{Multi-Faceted} \\ \textbf{Evaluation Metrics}} & 
        \begin{itemize}[leftmargin=0.4cm]
            \item \textbf{Video Physics Evaluation (VideoPhy) \cite{bansal2024videophy}:} Specifically assesses the physical plausibility of generated videos.
            \item \textbf{Video Score (VideoScore) \cite{he2024videoscore}:} Simulates fine-grained human feedback across multiple evaluation dimensions.
        \end{itemize} \\
        \bottomrule
    \end{tabular}
    }
    \caption{AIGVE-Tool Supported AIGVE Metrics Their Brief Introduction}
\end{table}


\newpage
\section{\ours Usage Paradigms}
\label{app:usage}

\ours is designed to facilitate AI-generated video evaluation with a configuration-driven and modular architecture. Built upon PyTorch~\cite{PyTorch} and MMEngine~\cite{mmengine2022}, \ours allowing users to seamlessly integrate new datasets, evaluation metrics, and experimental setups. This section provides a practical guide on using \ours for different evaluation paradigms.

\subsection{Configuration-Driven Execution}
\label{app-usage:config}

\ours leverages Python-based configuration files to define evaluation settings. Users can specify dataset sources, evaluation metrics, and model parameters without modifying core code. The example below illustrates how to configure an evaluation pipeline for the GSTVQA~\cite{GSTVQA} metric:

\begin{lstlisting}[language=Python, caption=Example configuration file for GSTVQA]
from mmengine.config import read_base
from mmengine.dataset import DefaultSampler
from metrics.video_quality_assessment.nn_based.gstvqa.gstvqa_metric import GSTVQA
from datasets import GSTVQADataset

with read_base():
    from ._base_.default import *

val_dataloader = dict(
    batch_size=1,
    num_workers=4,
    persistent_workers=True,
    drop_last=False,
    sampler=dict(type=DefaultSampler, shuffle=False),
    dataset=dict(
        type=GSTVQADataset,
        video_dir='/path/to/video_data/',
        prompt_dir='/path/to/annotations.json',
        model_name='vgg16',  # Supports 'vgg16' or 'resnet18'
        max_len=3,
    )
)

val_evaluator = dict(
    type=GSTVQA,
    model_path="metrics/video_quality_assessment/nn_based/gstvqa/GSTVQA/trained_model.pth",
)
\end{lstlisting}

This modular design allows users to switch datasets, evaluation models, and configurations by modifying only the config file instead of editing core framework components.

\subsection{Customizable Datasets}
\label{app-usage:dataset}

\ours provides a dataset abstraction, enabling support for various video datasets. Below is an example dataset implementation for GSTVQA~\cite{GSTVQA}, which extracts deep features from video frames using VGG16~\cite{VGG} or ResNet18~\cite{Resnet}.

\begin{lstlisting}[language=Python, caption=Example dataset class for GSTVQA]
import os
import torch
import cv2
import json
from torch.utils.data import Dataset
from torchvision import models
from core.registry import DATASETS

class FeatureExtractor(nn.Module):
    def __init__(self, model_name='vgg16'):
        super().__init__()
        if model_name == 'vgg16':
            model = models.vgg16(pretrained=True)
            self.feature_extractor = nn.Sequential(*list(model.features.children())[:-1])
        elif model_name == 'resnet18':
            model = models.resnet18(pretrained=True)
            self.feature_extractor = nn.Sequential(*list(model.children())[:-2])
        else:
            raise ValueError("Unsupported model: Choose 'vgg16' or 'resnet18'")
        self.feature_extractor.eval()

    def forward(self, x):
        x = self.feature_extractor(x)
        return x.mean(dim=(2, 3)), x.std(dim=(2, 3))  # Mean and Std pooling

@DATASETS.register_module()
class GSTVQADataset(Dataset):
    def __init__(self, video_dir, prompt_dir, model_name='vgg16', max_len=500):
        self.video_dir = video_dir
        self.prompt_dir = prompt_dir
        self.model_name = model_name
        self.feature_extractor = FeatureExtractor(model_name=model_name)

    def __getitem__(self, index):
        video_path = os.path.join(self.video_dir, self.video_list[index])
        frames = self._extract_frames(video_path)
        mean_features, std_features = self.feature_extractor(frames)
        deep_features = torch.cat((mean_features, std_features), dim=1)
        return deep_features, video_path
\end{lstlisting}

This implementation allows users to swap feature extractors and modify dataset preprocessing without changing the evaluation pipeline.

\subsection{Modular Metric}
\label{app-usage:metric}

Each metric in \ours is implemented as a standalone module, ensuring ease of integration and extensibility. Below is an example of the GSTVQA metric~\cite{GSTVQA}, which operates on extracted video features:

\begin{lstlisting}[language=Python, caption=Example metric class for GSTVQA]
import torch
import torch.nn as nn
from mmengine.evaluator import BaseMetric
from core.registry import METRICS

@METRICS.register_module()
class GSTVQA(BaseMetric):
    def __init__(self, model_path: str):
        super(GSTVQA, self).__init__()
        self.device = torch.device("cuda" if torch.cuda.is_available() else "cpu")
        self.model = self._load_model(model_path)

    def _load_model(self, model_path):
        model = nn.Sequential(nn.Linear(2944, 512), nn.ReLU(), nn.Linear(512, 1))
        model.load_state_dict(torch.load(model_path, map_location=self.device))
        model.eval()
        return model

    def process(self, data_batch, data_samples):
        deep_features, video_paths = data_samples
        with torch.no_grad():
            scores = self.model(deep_features)
        results = [{"video_path": video, "GSTVQA_Score": score.item()} for video, score in zip(video_paths, scores)]
        self.results.extend(results)

    def compute_metrics(self, results):
        avg_score = sum(r["GSTVQA_Score"] for r in results) / len(results)
        return {"GSTVQA_Mean_Score": avg_score}
\end{lstlisting}

This modular structure ensures that new evaluation metrics can be integrated seamlessly by following the standardized interface.

\subsection{Running \textbf{\ours}}

After defining the dataset and metrics in the configuration, users can run evaluations with a single command:

\begin{lstlisting}[language=bash, caption=Running AIGVE-Tool]
python tools/test.py configs/gstvqa.py --work-dir results/
\end{lstlisting}

This command executes the entire evaluation pipeline, automatically loading datasets, computing metrics, and generating structured reports.

\subsection{Summary}

\ours provides a flexible, standardized, and configurable framework for evaluating AI-generated videos. Its modular structure allows researchers to:

\begin{itemize}
    \item Define evaluation pipelines via config files.
    \item Use and customize datasets without modifying core logic.
    \item Integrate new metrics with minimal implementation effort.
    \item Run evaluations easily through a single command.
\end{itemize}

This paradigm significantly reduces the overhead for AI-generated video evaluation research, enabling rapid experimentation and extensible benchmarking.

%% file: tables/prompt_generation.tex
\begin{table*}[h]
    \centering
    \small
    \begin{tabular}{p{4cm} p{10cm}}
        \toprule
        \textbf{Prompt Type} & \textbf{Content} \\
        \midrule
        \texttt{system\_prompt} & "You are an expert in writing video generation prompts for video generation models." \\
        \midrule
        \texttt{global\_view\_prompt} & 
        Please help compose 10 video generation prompts that will be used to instruct video generation models to synthesize videos. In general, the generated videos should align with the real world. \\
        & \textbf{Guidelines:} \\
        & 1. The prompt should guide the model to generate realistic-style videos rather than 2D animation or stylistic art styles. \\
        & 2. The prompt should be 20-40 words long, describing the whole video in detail, including object attributes such as color and shape. \\
        & 3. The prompts should focus on a "global" view, such as a drone view of a city or a B-roll of a lake. The objects included must be: mountain, ocean, plain, river, lake, desert, city, architecture, or street. To enrich the prompt, consider variations in daylight, weather, and season or transitions between them. \\
        & 4. The video should be highly dynamic, reflected in camera movement or time transitions. \\
        & 5. The output should be a list of dictionaries, each containing three keys: \texttt{prompt} (generated prompt), \texttt{object} (included objects), and \texttt{dynamic} (types of motion, e.g., camera movement, daylight transition, weather transition, season transition). \\
        & 6. The prompts should be diverse and non-repetitive. \\
        \midrule
        \texttt{close\_shot\_prompt} & 
        Please help compose 10 video generation prompts that will be used to instruct video generation models to synthesize videos. In general, the generated videos should align with the real world. \\
        & \textbf{Guidelines:} \\
        & 1. The prompt should guide the model to generate realistic-style videos rather than 2D animation or stylistic art styles. \\
        & 2. The prompt should be 20-40 words long, describing the whole video in detail. \\
        & 3. The prompts should focus on a "close shot" of objects such as two people talking, water poured into a glass, or animals chasing each other. The subjects can be: human, animal, plant, or object. The prompt should include common objects found in daily life and describe attributes such as clothing colors or object shapes. \\
        & 4. The video should be highly dynamic. Dynamics can be single subject actions (e.g., walking), multiple subject interactions (e.g., two people talking), or camera movement. \\
        & 5. The output should be a list of dictionaries containing four keys: \texttt{prompt} (generated prompt), \texttt{subject} (included subjects), \texttt{dynamic} (e.g., single subject action, multiple subjects interaction, camera movement), and \texttt{detailed dynamic} (a general category of subject/camera action, such as conversation or chasing). \\
        & 6. The prompts should be diverse and non-repetitive. \\
        \bottomrule
    \end{tabular}
    \caption{The Prompt Used to Generate Video Instructions}
    \label{tab:video_prompts}
\end{table*}

%% file: tables/metrics.tex
\begin{table*}[h!]
    \centering
    \small
    \renewcommand{\arraystretch}{1.2} 
    \scalebox{0.95}{ 
    \begin{tabular}{l|p{13cm}}
        \toprule
        \textbf{Metric} & \textbf{Description} \\
        \midrule
        \textbf{Technical Quality} & Assesses the technical aspects of the video, including whether the resolution is sufficient for object recognition, whether the colors are natural, and whether there is an absence of noise or artifacts. \\
        \rowcolor[HTML]{F2F2F2}  \textbf{Dynamic} & Measures the extent of pixel changes throughout the video, focusing on significant object or camera movements and changes in environmental factors such as daylight, weather, or seasons. \\
        \textbf{Consistency} & Evaluates whether objects in the video maintain consistent properties, avoiding glitches, flickering, or unexpected changes. \\
        \rowcolor[HTML]{F2F2F2} \textbf{Physics} & Determines if the scene adheres to physical laws, ensuring that object behaviors and interactions are realistic and aligned with real-world physics. \\
        \textbf{Element Presence} & Checks if all objects mentioned in the instructions are present in the video. The score is based on the proportion of objects that are correctly included. \\
        \rowcolor[HTML]{F2F2F2} \textbf{Element Quality} & Assesses the realism and fidelity of objects in the video, awarding higher scores for detailed, natural, and visually appealing appearances. \\
        \textbf{Action/Interaction Presence} & Evaluates whether all actions and interactions described in the instructions are accurately represented in the video. \\
        \rowcolor[HTML]{F2F2F2} \textbf{Action/Interaction Quality} & Measures the naturalness and smoothness of actions and interactions, with higher scores for those that are realistic, lifelike, and seamlessly integrated into the scene. \\
        \textbf{Overall} & Reflects the comprehensive quality of the video based on all metrics, allowing raters to incorporate their subjective preferences into the evaluation. \\
        \bottomrule
    \end{tabular}
    }
    \caption{Metrics for Video Generation Evaluation}
    \label{tab:metrics}
\end{table*}

%% file: main.bbl
\begin{thebibliography}{10}\itemsep=-1pt

\bibitem{Bain21}
Max Bain, Arsha Nagrani, G{\"u}l Varol, and Andrew Zisserman.
\newblock Frozen in time: A joint video and image encoder for end-to-end retrieval.
\newblock In {\em IEEE International Conference on Computer Vision}, 2021.

\bibitem{bansal2024videophy}
Hritik Bansal, Zongyu Lin, Tianyi Xie, Zeshun Zong, Michal Yarom, Yonatan Bitton, Chenfanfu Jiang, Yizhou Sun, Kai-Wei Chang, and Aditya Grover.
\newblock Videophy: Evaluating physical commonsense for video generation.
\newblock {\em arXiv preprint arXiv:2406.03520}, 2024.

\bibitem{opencv}
G. Bradski.
\newblock {The OpenCV Library}.
\newblock {\em Dr. Dobb's Journal of Software Tools}, 2000.

\bibitem{GSTVQA}
Baoliang Chen, Lingyu Zhu, Guo Li, Hongfei Fan, and Shiqi Wang.
\newblock Learning generalized spatial-temporal deep feature representation for no-reference video quality assessment.
\newblock {\em arXiv preprint arXiv:2012.13936}, 2021.

\bibitem{MMDetection}
Kai Chen, Jiaqi Wang, Jiangmiao Pang, Yuhang Cao, Yu Xiong, Xiaoxiao Li, Shuyang Sun, Wansen Feng, Ziwei Liu, Jiarui Xu, Zheng Zhang, Dazhi Cheng, Chenchen Zhu, Tianheng Cheng, Qijie Zhao, Buyu Li, Xin Lu, Rui Zhu, Yue Wu, Jifeng Dai, Jingdong Wang, Jianping Shi, Wanli Ouyang, Chen~Change Loy, and Dahua Lin.
\newblock Mmdetection: Open mmlab detection toolbox and benchmark.
\newblock {\em arXiv preprint arXiv:1906.07155}, 2019.

\bibitem{DSG}
Jaemin Cho, Yushi Hu, Roopal Garg, Peter Anderson, Ranjay Krishna, Jason Baldridge, Mohit Bansal, Jordi Pont-Tuset, and Su Wang.
\newblock Davidsonian scene graph: Improving reliability in fine-grained evaluation for text-to-image generation.
\newblock {\em arXiv preprint arXiv:2310.18235}, 2024.

\bibitem{Keras}
Fran\c{c}ois Chollet et~al.
\newblock Keras.
\newblock \url{https://github.com/fchollet/keras}, 2015.

\bibitem{MMDetection3D}
MMDetection3D Contributors.
\newblock {MMDetection3D: OpenMMLab} next-generation platform for general {3D} object detection.
\newblock \url{https://github.com/open-mmlab/mmdetection3d}, 2020.

\bibitem{MMSegmentation}
MMSegmentation Contributors.
\newblock {MMSegmentation}: Openmmlab semantic segmentation toolbox and benchmark.
\newblock \url{https://github.com/open-mmlab/mmsegmentation}, 2020.

\bibitem{MMPose}
MMPose Contributors.
\newblock Openmmlab pose estimation toolbox and benchmark.
\newblock \url{https://github.com/open-mmlab/mmpose}, 2020.

\bibitem{mmengine2022}
MMEngine Contributors.
\newblock {MMEngine}: Openmmlab foundational library for training deep learning models.
\newblock \url{https://github.com/open-mmlab/mmengine}, 2022.

\bibitem{AllenNLP}
Matt Gardner, Joel Grus, Mark Neumann, Oyvind Tafjord, Pradeep Dasigi, Nelson~F. Liu, Matthew Peters, Michael Schmitz, and Luke~S. Zettlemoyer.
\newblock Allennlp: A deep semantic natural language processing platform.
\newblock {\em arXiv preprint arXiv:1803.07640}, 2017.

\bibitem{Resnet}
Kaiming He, Xiangyu Zhang, Shaoqing Ren, and Jian Sun.
\newblock Deep residual learning for image recognition.
\newblock In {\em 2016 IEEE Conference on Computer Vision and Pattern Recognition (CVPR)}, pages 770--778, 2016.

\bibitem{he2024videoscore}
Xuan He, Dongfu Jiang, Ge Zhang, Max Ku, Achint Soni, Sherman Siu, Haonan Chen, Abhranil Chandra, Ziyan Jiang, Aaran Arulraj, Kai Wang, Quy~Duc Do, Yuansheng Ni, Bohan Lyu, Yaswanth Narsupalli, Rongqi Fan, Zhiheng Lyu, Yuchen Lin, and Wenhu Chen.
\newblock Videoscore: Building automatic metrics to simulate fine-grained human feedback for video generation.
\newblock In {\em Proceedings of the Empirical Methods in Natural Language Processing (EMNLP)}, pages 2105--2123. Association for Computational Linguistics, 2024.

\bibitem{hessel2022clipscore}
Jack Hessel, Ari Holtzman, Maxwell Forbes, Ronan~Le Bras, and Yejin Choi.
\newblock Clipscore: A reference-free evaluation metric for image captioning.
\newblock In {\em Proceedings of the Empirical Methods in Natural Language Processing (EMNLP)}, 2021.

\bibitem{FID}
Martin Heusel, Hubert Ramsauer, Thomas Unterthiner, Bernhard Nessler, and Sepp Hochreiter.
\newblock Gans trained by a two time-scale update rule converge to a local nash equilibrium.
\newblock In {\em Advances in Neural Information Processing Systems (NeurIPS)}, 2017.

\bibitem{ho2020denoising}
Jonathan Ho, Ajay Jain, and Pieter Abbeel.
\newblock Denoising diffusion probabilistic models.
\newblock In {\em Advances in Neural Information Processing Systems (NeurIPS)}, 2020.

\bibitem{konvid1k}
Vlad Hosu, Franz Hahn, Mohsen Jenadeleh, Hanhe Lin, Hui Men, Tam{\'a}s Szir{\'a}nyi, Shujun Li, and Dietmar Saupe.
\newblock The konstanz natural video database.
\newblock In {\em 2017 Ninth International Conference on Quality of Multimedia Experience (QoMEX)}, 2017.

\bibitem{hu2023tifa}
Yushi Hu, Benlin Liu, Jungo Kasai, Yizhong Wang, Mari Ostendorf, Ranjay Krishna, and Noah~A Smith.
\newblock Tifa: Accurate and interpretable text-to-image faithfulness evaluation with question answering.
\newblock In {\em IEEE/CVF International Conference on Computer Vision (ICCV)}, 2023.

\bibitem{huang_vbench_2023}
Ziqi Huang, Yinan He, Jiashuo Yu, Fan Zhang, Chenyang Si, Yuming Jiang, Yuanhan Zhang, Tianxing Wu, Qingyang Jin, Nattapol Chanpaisit, Yaohui Wang, Xinyuan Chen, Limin Wang, Dahua Lin, Yu Qiao, and Ziwei Liu.
\newblock {VBench}: {Comprehensive} {Benchmark} {Suite} for {Video} {Generative} {Models}.
\newblock In {\em IEEE/CVF Conference on Computer Vision and Pattern Recognition (CVPR)}, 2024.

\bibitem{jin2024pyramidal}
Yang Jin, Zhicheng Sun, Ningyuan Li, Kun Xu, Kun Xu, Hao Jiang, Nan Zhuang, Quzhe Huang, Yang Song, Yadong Mu, and Zhouchen Lin.
\newblock Pyramidal flow matching for efficient video generative modeling.
\newblock In {\em Proceedings of the International Conference on Learning Representations (ICLR)}, 2025.

\bibitem{PickScore}
Yuval Kirstain, Adam Polyak, Uriel Singer, Shahbuland Matiana, Joe Penna, and Omer Levy.
\newblock Pick-a-pic: An open dataset of user preferences for text-to-image generation.
\newblock In {\em Neural Information Processing Systems (NeurIPS)}, 2023.

\bibitem{kong2025hunyuanvideosystematicframeworklarge}
Weijie Kong, Qi Tian, Zijian Zhang, Rox Min, Zuozhuo Dai, Jin Zhou, Jiangfeng Xiong, Xin Li, Bo Wu, Jianwei Zhang, Kathrina Wu, Qin Lin, Junkun Yuan, Yanxin Long, Aladdin Wang, Andong Wang, Changlin Li, Duojun Huang, Fang Yang, Hao Tan, Hongmei Wang, Jacob Song, Jiawang Bai, Jianbing Wu, Jinbao Xue, Joey Wang, Kai Wang, Mengyang Liu, Pengyu Li, Shuai Li, Weiyan Wang, Wenqing Yu, Xinchi Deng, Yang Li, Yi Chen, Yutao Cui, Yuanbo Peng, Zhentao Yu, Zhiyu He, Zhiyong Xu, Zixiang Zhou, Zunnan Xu, Yangyu Tao, Qinglin Lu, Songtao Liu, Dax Zhou, Hongfa Wang, Yong Yang, Di Wang, Yuhong Liu, Jie Jiang, and Caesar Zhong.
\newblock Hunyuanvideo: A systematic framework for large video generative models.
\newblock {\em arXiv preprint arXiv:2412.03603}, 2024.

\bibitem{ku2024viescore}
Max Ku, Dongfu Jiang, Cong Wei, Xiang Yue, and Wenhu Chen.
\newblock Viescore: Towards explainable metrics for conditional image synthesis evaluation.
\newblock In {\em Proceedings of the 62nd Annual Meeting of the Association for Computational Linguistics}. Association for Computational Linguistics, 2024.

\bibitem{li2023blip}
Junnan Li, Dongxu Li, Silvio Savarese, and Steven Hoi.
\newblock Blip-2: Bootstrapping language-image pre-training with frozen image encoders and large language models.
\newblock In {\em Proceedings of the 41st International Conference on Machine Learning}, pages 19730--19742, 2023.

\bibitem{liao2024evaluation}
Mingxiang Liao, Hannan Lu, Xinyu Zhang, Fang Wan, Tianyu Wang, Yuzhong Zhao, Wangmeng Zuo, Qixiang Ye, and Jingdong Wang.
\newblock Evaluation of text-to-video generation models: A dynamics perspective.
\newblock {\em arXiv preprint arXiv:2407.01094}, 2024.

\bibitem{liu2024surveyaigve}
Xiao Liu, Xinhao Xiang, Zizhong Li, Yongheng Wang, Zhuoheng Li, Zhuosheng Liu, Weidi Zhang, Weiqi Ye, and Jiawei Zhang.
\newblock A survey of ai-generated video evaluation.
\newblock {\em arXiv preprint arXiv:2410.19884}, 2024.

\bibitem{liu2023evalcrafter}
Yaofang Liu, Xiaodong Cun, Xuebo Liu, Xintao Wang, Yong Zhang, Haoxin Chen, Yang Liu, Tieyong Zeng, Raymond Chan, and Ying Shan.
\newblock Evalcrafter: Benchmarking and evaluating large video generation models.
\newblock In {\em IEEE/CVF Conference on Computer Vision and Pattern Recognition (CVPR)}, 2024.

\bibitem{NIQE}
Anish Mittal, Anush~Krishna Moorthy, and Alan~Conrad Bovik.
\newblock No-reference image quality assessment in the spatial domain.
\newblock {\em IEEE Transactions on Image Processing}, 21(12):4695--4708, 2012.

\bibitem{CVD2014}
Mikko Nuutinen, Toni Virtanen, Mikko Vaahteranoksa, Tero Vuori, Pirkko Oittinen, and Jukka Häkkinen.
\newblock Cvd2014—a database for evaluating no-reference video quality assessment algorithms.
\newblock {\em IEEE Transactions on Image Processing}, 25(7):3073--3086, 2016.

\bibitem{sora2024}
OpenAI.
\newblock Sora.
\newblock \url{https://openai.com/index/sora/}, 2024.

\bibitem{openai2024gpt4technicalreport}
OpenAI, Josh Achiam, Steven Adler, Sandhini Agarwal, Lama Ahmad, Ilge Akkaya, Florencia~Leoni Aleman, Diogo Almeida, Janko Altenschmidt, Sam Altman, Shyamal Anadkat, Red Avila, Igor Babuschkin, Suchir Balaji, Valerie Balcom, Paul Baltescu, Haiming Bao, Mohammad Bavarian, Jeff Belgum, Irwan Bello, Jake Berdine, Gabriel Bernadett-Shapiro, Christopher Berner, Lenny Bogdonoff, Oleg Boiko, Madelaine Boyd, Anna-Luisa Brakman, Greg Brockman, Tim Brooks, Miles Brundage, Kevin Button, Trevor Cai, Rosie Campbell, Andrew Cann, Brittany Carey, Chelsea Carlson, Rory Carmichael, Brooke Chan, Che Chang, Fotis Chantzis, Derek Chen, Sully Chen, Ruby Chen, Jason Chen, Mark Chen, Ben Chess, Chester Cho, Casey Chu, Hyung~Won Chung, Dave Cummings, Jeremiah Currier, Yunxing Dai, Cory Decareaux, Thomas Degry, Noah Deutsch, Damien Deville, Arka Dhar, David Dohan, Steve Dowling, Sheila Dunning, Adrien Ecoffet, Atty Eleti, Tyna Eloundou, David Farhi, Liam Fedus, Niko Felix, Simón~Posada Fishman, Juston Forte, Isabella Fulford, Leo
  Gao, Elie Georges, Christian Gibson, Vik Goel, Tarun Gogineni, Gabriel Goh, Rapha Gontijo-Lopes, Jonathan Gordon, Morgan Grafstein, Scott Gray, Ryan Greene, Joshua Gross, Shixiang~Shane Gu, Yufei Guo, Chris Hallacy, Jesse Han, Jeff Harris, Yuchen He, Mike Heaton, Johannes Heidecke, Chris Hesse, Alan Hickey, Wade Hickey, Peter Hoeschele, Brandon Houghton, Kenny Hsu, Shengli Hu, Xin Hu, Joost Huizinga, Shantanu Jain, Shawn Jain, Joanne Jang, Angela Jiang, Roger Jiang, Haozhun Jin, Denny Jin, Shino Jomoto, Billie Jonn, Heewoo Jun, Tomer Kaftan, Łukasz Kaiser, Ali Kamali, Ingmar Kanitscheider, Nitish~Shirish Keskar, Tabarak Khan, Logan Kilpatrick, Jong~Wook Kim, Christina Kim, Yongjik Kim, Jan~Hendrik Kirchner, Jamie Kiros, Matt Knight, Daniel Kokotajlo, Łukasz Kondraciuk, Andrew Kondrich, Aris Konstantinidis, Kyle Kosic, Gretchen Krueger, Vishal Kuo, Michael Lampe, Ikai Lan, Teddy Lee, Jan Leike, Jade Leung, Daniel Levy, Chak~Ming Li, Rachel Lim, Molly Lin, Stephanie Lin, Mateusz Litwin, Theresa Lopez, Ryan
  Lowe, Patricia Lue, Anna Makanju, Kim Malfacini, Sam Manning, Todor Markov, Yaniv Markovski, Bianca Martin, Katie Mayer, Andrew Mayne, Bob McGrew, Scott~Mayer McKinney, Christine McLeavey, Paul McMillan, Jake McNeil, David Medina, Aalok Mehta, Jacob Menick, Luke Metz, Andrey Mishchenko, Pamela Mishkin, Vinnie Monaco, Evan Morikawa, Daniel Mossing, Tong Mu, Mira Murati, Oleg Murk, David Mély, Ashvin Nair, Reiichiro Nakano, Rajeev Nayak, Arvind Neelakantan, Richard Ngo, Hyeonwoo Noh, Long Ouyang, Cullen O'Keefe, Jakub Pachocki, Alex Paino, Joe Palermo, Ashley Pantuliano, Giambattista Parascandolo, Joel Parish, Emy Parparita, Alex Passos, Mikhail Pavlov, Andrew Peng, Adam Perelman, Filipe de Avila Belbute~Peres, Michael Petrov, Henrique~Ponde de Oliveira~Pinto, Michael, Pokorny, Michelle Pokrass, Vitchyr~H. Pong, Tolly Powell, Alethea Power, Boris Power, Elizabeth Proehl, Raul Puri, Alec Radford, Jack Rae, Aditya Ramesh, Cameron Raymond, Francis Real, Kendra Rimbach, Carl Ross, Bob Rotsted, Henri Roussez,
  Nick Ryder, Mario Saltarelli, Ted Sanders, Shibani Santurkar, Girish Sastry, Heather Schmidt, David Schnurr, John Schulman, Daniel Selsam, Kyla Sheppard, Toki Sherbakov, Jessica Shieh, Sarah Shoker, Pranav Shyam, Szymon Sidor, Eric Sigler, Maddie Simens, Jordan Sitkin, Katarina Slama, Ian Sohl, Benjamin Sokolowsky, Yang Song, Natalie Staudacher, Felipe~Petroski Such, Natalie Summers, Ilya Sutskever, Jie Tang, Nikolas Tezak, Madeleine~B. Thompson, Phil Tillet, Amin Tootoonchian, Elizabeth Tseng, Preston Tuggle, Nick Turley, Jerry Tworek, Juan Felipe~Cerón Uribe, Andrea Vallone, Arun Vijayvergiya, Chelsea Voss, Carroll Wainwright, Justin~Jay Wang, Alvin Wang, Ben Wang, Jonathan Ward, Jason Wei, CJ Weinmann, Akila Welihinda, Peter Welinder, Jiayi Weng, Lilian Weng, Matt Wiethoff, Dave Willner, Clemens Winter, Samuel Wolrich, Hannah Wong, Lauren Workman, Sherwin Wu, Jeff Wu, Michael Wu, Kai Xiao, Tao Xu, Sarah Yoo, Kevin Yu, Qiming Yuan, Wojciech Zaremba, Rowan Zellers, Chong Zhang, Marvin Zhang, Shengjia
  Zhao, Tianhao Zheng, Juntang Zhuang, William Zhuk, and Barret Zoph.
\newblock Gpt-4 technical report, 2024.

\bibitem{PyTorch}
Adam Paszke, Sam Gross, Francisco Massa, Adam Lerer, James Bradbury, Gregory Chanan, Trevor Killeen, Zeming Lin, Natalia Gimelshein, Luca Antiga, Alban Desmaison, Andreas Köpf, Edward Yang, Zach DeVito, Martin Raison, Alykhan Tejani, Sasank Chilamkurthy, Benoit Steiner, Lu Fang, Junjie Bai, and Soumith Chintala.
\newblock Pytorch: An imperative style, high-performance deep learning library, 2019.

\bibitem{radford2021learningtransferablevisualmodels}
Alec Radford, Jong~Wook Kim, Chris Hallacy, Aditya Ramesh, Gabriel Goh, Sandhini Agarwal, Girish Sastry, Amanda Askell, Pamela Mishkin, Jack Clark, Gretchen Krueger, and Ilya Sutskever.
\newblock Learning transferable visual models from natural language supervision.
\newblock In {\em Proceedings of the 38th International Conference on Machine Learning}. PMLR, 2021.

\bibitem{IS_Score}
Tim Salimans, Ian Goodfellow, Wojciech Zaremba, Vicki Cheung, Alec Radford, and Xi Chen.
\newblock Improved techniques for training gans.
\newblock In {\em Advances in Neural Information Processing Systems (NeurIPS)}, 2016.

\bibitem{VGG}
Karen Simonyan and Andrew Zisserman.
\newblock Very deep convolutional networks for large-scale image recognition.
\newblock In {\em 3rd International Conference on Learning Representations (ICLR)}, 2015.

\bibitem{LiveVQC}
Zeina Sinno and Alan~Conrad Bovik.
\newblock Large-scale study of perceptual video quality.
\newblock {\em IEEE Transactions on Image Processing}, 28(2):612--627, 2018.

\bibitem{SimpleVQA}
Wei Sun, Xiongkuo Min, Wei Lu, and Guangtao Zhai.
\newblock A deep learning based no-reference quality assessment model for ugc videos.
\newblock In {\em Proceedings of the 30th ACM International Conference on Multimedia}, MM ’22, page 856–865. ACM, Oct. 2022.

\bibitem{genmo2024mochi}
Genmo Team.
\newblock Mochi 1.
\newblock \url{https://github.com/genmoai/models}, 2024.

\bibitem{OpenPCDet}
OpenPCDet~Development Team.
\newblock Openpcdet: An open-source toolbox for 3d object detection from point clouds.
\newblock \url{https://github.com/open-mmlab/OpenPCDet}, 2020.

\bibitem{MuJoCo}
Emanuel Todorov, Tom Erez, and Yuval Tassa.
\newblock Mujoco: A physics engine for model-based control.
\newblock In {\em 2012 IEEE/RSJ International Conference on Intelligent Robots and Systems}, pages 5026--5033. IEEE, 2012.

\bibitem{FVD}
Thomas Unterthiner, Sjoerd van Steenkiste, Karol Kurach, Raphael Marinier, Marcin Michalski, and Sylvain Gelly.
\newblock Towards accurate generative models of video: A new metric \& challenges.
\newblock {\em arXiv preprint arXiv:1812.01717}, 2019.

\bibitem{Wang2004ImageQA}
Zhou Wang, Alan~Conrad Bovik, Hamid~R. Sheikh, and Eero~P. Simoncelli.
\newblock Image quality assessment: from error visibility to structural similarity.
\newblock {\em IEEE Transactions on Image Processing}, 13:600--612, 2004.

\bibitem{WANG2004121}
Zhou Wang, Ligang Lu, and Alan~C. Bovik.
\newblock Video quality assessment based on structural distortion measurement.
\newblock {\em Signal Processing: Image Communication}, 19(2):121--132, 2004.

\bibitem{ESPnet}
Shinji Watanabe, Takaaki Hori, Shigeki Karita, Tomoki Hayashi, Jiro Nishitoba, Yuya Unno, Nelson {Enrique Yalta Soplin}, Jahn Heymann, Matthew Wiesner, Nanxin Chen, Adithya Renduchintala, and Tsubasa Ochiai.
\newblock {ESPnet}: End-to-end speech processing toolkit.
\newblock In {\em Proceedings of Interspeech}, pages 2207--2211, 2018.

\bibitem{HuggingFace}
Thomas Wolf, Lysandre Debut, Victor Sanh, Julien Chaumond, Clement Delangue, Anthony Moi, Pierric Cistac, Tim Rault, Rémi Louf, Morgan Funtowicz, Joe Davison, Sam Shleifer, Patrick von Platen, Clara Ma, Yacine Jernite, Julien Plu, Canwen Xu, Teven~Le Scao, Sylvain Gugger, Mariama Drame, Quentin Lhoest, and Alexander~M. Rush.
\newblock Huggingface's transformers: State-of-the-art natural language processing, 2020.

\bibitem{Detectron2}
Yuxin Wu, Alexander Kirillov, Francisco Massa, Wan-Yen Lo, and Ross Girshick.
\newblock Detectron2.
\newblock \url{https://github.com/facebookresearch/detectron2}, 2019.

\bibitem{xing2024survey}
Zhen Xing, Qijun Feng, Haoran Chen, Qi Dai, Han Hu, Hang Xu, Zuxuan Wu, and Yu-Gang Jiang.
\newblock A survey on video diffusion models.
\newblock {\em ACM Computing Surveys}, 57(2):1--42, 2024.

\bibitem{xu2016msr}
Jun Xu, Tao Mei, Ting Yao, and Yong Rui.
\newblock Msr-vtt: A large video description dataset for bridging video and language.
\newblock In {\em IEEE/CVF Conference on Computer Vision and Pattern Recognition (CVPR)}, pages 5288--5296, 2016.

\bibitem{Hydra}
Omry Yadan.
\newblock Hydra - a framework for elegantly configuring complex applications.
\newblock Github, 2019.

\bibitem{yang2024cogvideox}
Zhuoyi Yang, Jiayan Teng, Wendi Zheng, Ming Ding, Shiyu Huang, Jiazheng Xu, Yuanming Yang, Wenyi Hong, Xiaohan Zhang, Guanyu Feng, et~al.
\newblock Cogvideox: Text-to-video diffusion models with an expert transformer.
\newblock {\em arXiv preprint arXiv:2408.06072}, 2024.

\bibitem{UGVQ}
Zhichao Zhang, Xinyue Li, Wei Sun, Jun Jia, Xiongkuo Min, Zicheng Zhang, Chunyi Li, Zijian Chen, Puyi Wang, Zhongpeng Ji, Fengyu Sun, Shangling Jui, and Guangtao Zhai.
\newblock Benchmarking aigc video quality assessment: A dataset and unified model.
\newblock {\em arXiv preprint arXiv:2407.21408}, 2024.

\bibitem{Light_VQA_plus}
Xunchu Zhou, Xiaohong Liu, Yunlong Dong, Tengchuan Kou, Yixuan Gao, Zicheng Zhang, Chunyi Li, Haoning Wu, and Guangtao Zhai.
\newblock Light-vqa+: A video quality assessment model for exposure correction with vision-language guidance.
\newblock {\em arXiv preprint arXiv:2405.03333}, 2024.

\end{thebibliography}
